\documentclass[10pt,aps,prd,email,showpacs,preprintnumbers,showkeys,superscriptaddress,amsmath,amssymb]{revtex4-2}
\usepackage{pgfplots}
\usepackage{tikz}
\usetikzlibrary{positioning}
\usepackage{color} 
\usepackage{hyperref}
\usepackage{slashed}
\usepackage{graphicx}
\usepackage{latexsym}
\usepackage{epstopdf}
\usepackage{amsmath}
\usepackage{enumitem}
\usepackage{amssymb,amsmath}
\usepackage{multirow}
\usepackage{mathtools}
\usepackage{bbm}
\usepackage{bm}
\usepackage{amsfonts}
\usepackage{amssymb}
\usepackage{tabularx}
\usepackage{calligra}
\usepackage{calrsfs}
\usepackage{makecell}
\usepackage{placeins}   
\usepackage{float}      
\usepackage{dblfloatfix} 
\usepackage[normalem]{ulem}
\usepackage[USenglish,american]{babel}
\usepackage{xcolor}
\usepackage{hyperref}
\hypersetup{
  colorlinks=false,             
  linkbordercolor=blue,         
  citebordercolor=green,
  urlbordercolor=blue,
  pdfborder={0 0 1}             
}
\usepackage{subcaption}
\usepackage[compat=1.1.0]{tikz-feynman}

\expandafter\ifx\csname package@font\endcsname\relax\else
 \expandafter\expandafter
 \expandafter\usepackage
 \expandafter\expandafter
 \expandafter{\csname package@font\endcsname}%
\fi
\begin{document}
\title{Scale redundancy and soft gauge fixing in positively homogeneous neural networks}

\author{Rodrigo Carmo Terin} 
\email{rodrigo.carmo@urjc.es}
\affiliation{King Juan Carlos University, Facuty of Experimental Sciences and Technology, Department of Applied Physics, Av. del Alcalde de Móstoles, 28933, Madrid, Spain}

\begin{abstract}
Neural networks with positively homogeneous activations exhibit an exact
continuous reparametrization symmetry: neuron-wise rescalings generate
parameter-space orbits along which the input--output function is invariant.
We interpret this symmetry as a gauge redundancy and introduce gauge-adapted
coordinates that separate invariant and scale-imbalance directions.
Inspired by gauge fixing in field theory, we introduce a soft orbit-selection
(norm-balancing) functional acting only on redundant scale coordinates.
We show analytically that it induces dissipative relaxation of imbalance modes
to preserve the realized function.
In controlled experiments, this orbit-selection penalty expands the stable
learning-rate regime and suppresses scale drift without changing expressivity.
These results establish a structural link between gauge-orbit geometry
and optimization conditioning, providing a concrete connection between
gauge-theoretic concepts and machine learning.
\end{abstract}

\maketitle

\section{Introduction}
\label{sec:introduction}

Deep neural networks (DNNs) achieve remarkable empirical performance,
yet the geometry of their parameter space remains only partially understood.
In particular, optimization stability can depend strongly on learning rate,
initialization scale, and internal parametrization choices,
even when the realized input--output function is unchanged.
This observation has motivated a broad effort to import ideas from
statistical physics and field theory into deep learning,
including large-$N$ expansions, diagrammatic techniques,
and effective descriptions near criticality
\cite{Yaida2020,DyerGurAri2020,BondesanWelling2021,
HalversonMaitiStoner2021,ErbinLahocheSamary2022,Roberts2022,
GrosvenorJefferson2022,CarmoTerin:2025lyx}.
In particular, stability near the edge of chaos has been shown to be governed by structural constraints rather than task-specific details
\cite{Sompolinsky1988,poole2016exponential,Langton1990,Roberts2022,CarmoTerin:2025lyx}.

A fundamental structural feature of many modern architectures is
positive homogeneity of activation functions such as ReLU.
Positive homogeneity implies that neuron-wise rescalings of incoming and outgoing weights can leave the network function invariant.
This rescaling invariance is well known and has been exploited in several contexts.
Scale-invariant optimization methods such as G-SGD explicitly operate in quotient-like parameter spaces induced by this symmetry
\cite{Meng2019GSGD},
while weight-balancing procedures provide function-preserving repair flows that reduce internal imbalance without altering the realized mapping
\cite{Saul2023WeightBalancing}.
More generally, recent analyses have asized that parameter-space symmetries give rise to degeneracies and flat directions in the loss landscape
\cite{Brea2019WeightSymmetry}.

The present work does not rediscover this invariance.
Instead, we make its orbit structure explicit and treat neuron-wise rescaling
as a gauge redundancy in parameter space.
Under this viewpoint, parameter configurations related by positive diagonal rescalings form continuous equivalence classes (orbits)
that represent the same function.
The task loss is constant along these directions,
implying the existence of flat gauge modes that are invisible at the level of the input--output mapping.

Neuron-wise scale invariance induced by positive homogeneity has been previously
studied from multiple perspectives.
Scale-invariant optimization methods such as G-SGD explicitly perform updates
in positively scale-invariant spaces \cite{Meng2019GSGD},
while weight-balancing procedures construct function-preserving flows
that reduce internal imbalance \cite{Saul2023WeightBalancing}.
Permutation symmetries and other parameter-space degeneracies have also been
shown to induce flat valleys and saddle structures in loss landscapes
\cite{Brea2019WeightSymmetry}.

Moreover our approach differs in emphasis and construction.
Rather than reparametrizing the optimization algorithm or applying explicit repair steps,
we introduce a variational gauge-fixing functional that acts exclusively on
the redundant scale coordinates while leaving gauge-invariant quantities untouched.
This provides (i) an explicit orbit–slice decomposition of parameter space,
(ii) a tunable conditioning mechanism integrated into the loss,
and (iii) a diagnostic framework based on gauge coordinates that
quantifies scale drift and stability boundaries.
In this sense, the gauge-theoretic formulation serves not merely as terminology,
but as a geometric principle organizing the role of reparametrization redundancy
in optimization.

Interpreting rescaling invariance in orbit language enables us to introduce
a simple variational gauge-fixing principle.
We define explicit {gauge coordinates} associated with neuron-wise scale imbalance
and construct a soft gauge-fixing functional that acts only on these coordinates,
leaving orbit invariants untouched.
Unlike global regularizers such as weight decay,
the proposed term does not penalize overall magnitude,
and unlike batch normalization it does not modify the forward pass.
It instead selects a canonical representative within each rescaling orbit.
Therefore,
our contribution is threefold:
\begin{enumerate}
\item[(i)] We formalize neuron-wise rescaling as a continuous orbit structure
in parameter space and identify explicit gauge coordinates
that separate invariant and redundant directions.

\item[(ii)] We introduce a soft gauge-fixing functional
that acts exclusively on the redundant scale coordinates
and prove the existence of a balanced representative on each orbit.

\item[(iii)] We quantify the geometric impact of lifting these flat directions
through stability-region diagnostics and scale-drift measurements,
showing that gauge fixing expands the empirically stable learning-rate regime
while preserving the realized function up to machine precision.
\end{enumerate}

The emphasis of this work is therefore geometric rather than purely empirical.
We do not claim large improvements in validation error in minimal tasks.
Instead, we demonstrate that redundant rescaling directions
modify the effective conditioning encountered by gradient descent,
and that softly lifting them alters the stability boundary of optimization.
This perspective complements existing work on scale-invariant optimization
and weight-balancing flows.
Whereas prior approaches either repair imbalance through explicit function-preserving updates
\cite{Saul2023WeightBalancing}
or reparametrize optimization in quotient spaces
\cite{Meng2019GSGD},
we introduce a tunable variational gauge-fixing term
that can be integrated directly into the training objective
and whose effect can be diagnosed through orbit-based coordinates.
In this sense, the gauge interpretation provides both a unifying geometric language
and a practical conditioning mechanism.

The paper is organized as follows.
Section~\ref{sec:gauge_symmetry} formalizes neuron-wise rescaling as a gauge redundancy
and introduces gauge-adapted coordinates that separate invariant and redundant directions.
Section~\ref{sec:gauge_fixing} presents the norm-balancing functional
and proves the existence of balanced representatives on each gauge orbit.
Section~\ref{sec:gauge_flow} derives the gradient-flow dynamics in gauge coordinates,
showing that the proposed gauge-fixing term induces an explicit dissipative relaxation
of the imbalance modes.
Section~\ref{sec:experiments} provides stability-region diagnostics,
learning-rate stress tests, and scale-drift measurements validating the theoretical analysis.
We conclude in Sec.~\ref{sec:conclusion}
with implications for deeper homogeneous architectures and optimization geometry. Detailed derivations of the gauge-coordinate gradients and the resulting flow equations are collected in
App.~\ref{app:gauge_flow_details}.

\section{Reparametrization Symmetry as Gauge Redundancy}
\label{sec:gauge_symmetry}

We begin by identifying an exact continuous reparametrization symmetry in
neural networks with positively homogeneous activation functions and
interpreting it as a gauge redundancy in parameter space.

Consider a fully connected neural network with a single hidden layer and ReLU activation,
\begin{equation}
f(x) = W_2 \, \sigma(W_1 x + b_1) + b_2 ,
\label{eq:relu_net}
\end{equation}
where $\sigma(z)=\max(0,z)$ is applied componentwise,
$W_1 \in \mathbb{R}^{H \times d}$,
$b_1 \in \mathbb{R}^{H}$,
$W_2 \in \mathbb{R}^{m \times H}$,
and $b_2 \in \mathbb{R}^{m}$.
In the experiments of Sec.~\ref{sec:experiments} we consider the scalar-output case $m=1$,
but the symmetry structure discussed below holds more generally.

The ReLU function is positively homogeneous of degree one,
\begin{equation}
\sigma(\alpha z) = \alpha \sigma(z), \qquad \alpha > 0 .
\end{equation}
Let $D=\mathrm{diag}(s_1,\ldots,s_H)$ be a diagonal matrix with strictly positive entries.
Define the parameter transformation
\begin{equation}
\begin{aligned}
W_1 &\;\to\; D W_1, \qquad
b_1 \;\to\; D b_1, \\
W_2 &\;\to\; W_2 D^{-1}, \qquad
b_2 \;\to\; b_2 .
\end{aligned}
\label{eq:gauge_transform}
\end{equation}
Using the homogeneity of the activation, one immediately finds
\begin{equation}
\begin{aligned}
f(x) \;\to\;&\; W_2 D^{-1} \, \sigma\!\big(D (W_1 x + b_1)\big) + b_2 \\
=\;&\; W_2 \, \sigma(W_1 x + b_1) + b_2 .
\end{aligned}
\end{equation}
so that the input--output mapping of the network is exactly invariant under
the transformation~\eqref{eq:gauge_transform}.
Equation~\eqref{eq:gauge_transform} therefore generates a continuous family
of parameter configurations that all represent the same function $f(x)$.
The positive variables $\{s_i\}$ label equivalence classes of
physically indistinguishable parameterizations.
Importantly, this symmetry does not arise from invariance of the data or of the task,
but from an internal redundancy of the parameter space.

The structure is directly analogous to gauge symmetry in field theory.
In gauge theories, multiple field configurations related by local
transformations correspond to the same physical state.
Here, the redundancy acts in parameter space rather than spacetime,
but the mathematical structure is equivalent:
the loss function is invariant along continuous orbits generated
by~\eqref{eq:gauge_transform}.
From the perspective of optimization, this symmetry implies the existence of
flat directions in parameter space.
Along these directions, the loss remains unchanged while the norms of
individual parameter blocks can vary arbitrarily.
Although such degeneracy does not alter the expressive power of the network,
it allows large internal rescalings that can influence optimization dynamics.

\begin{figure}[t]
\centering
\includegraphics[width=0.58\linewidth]{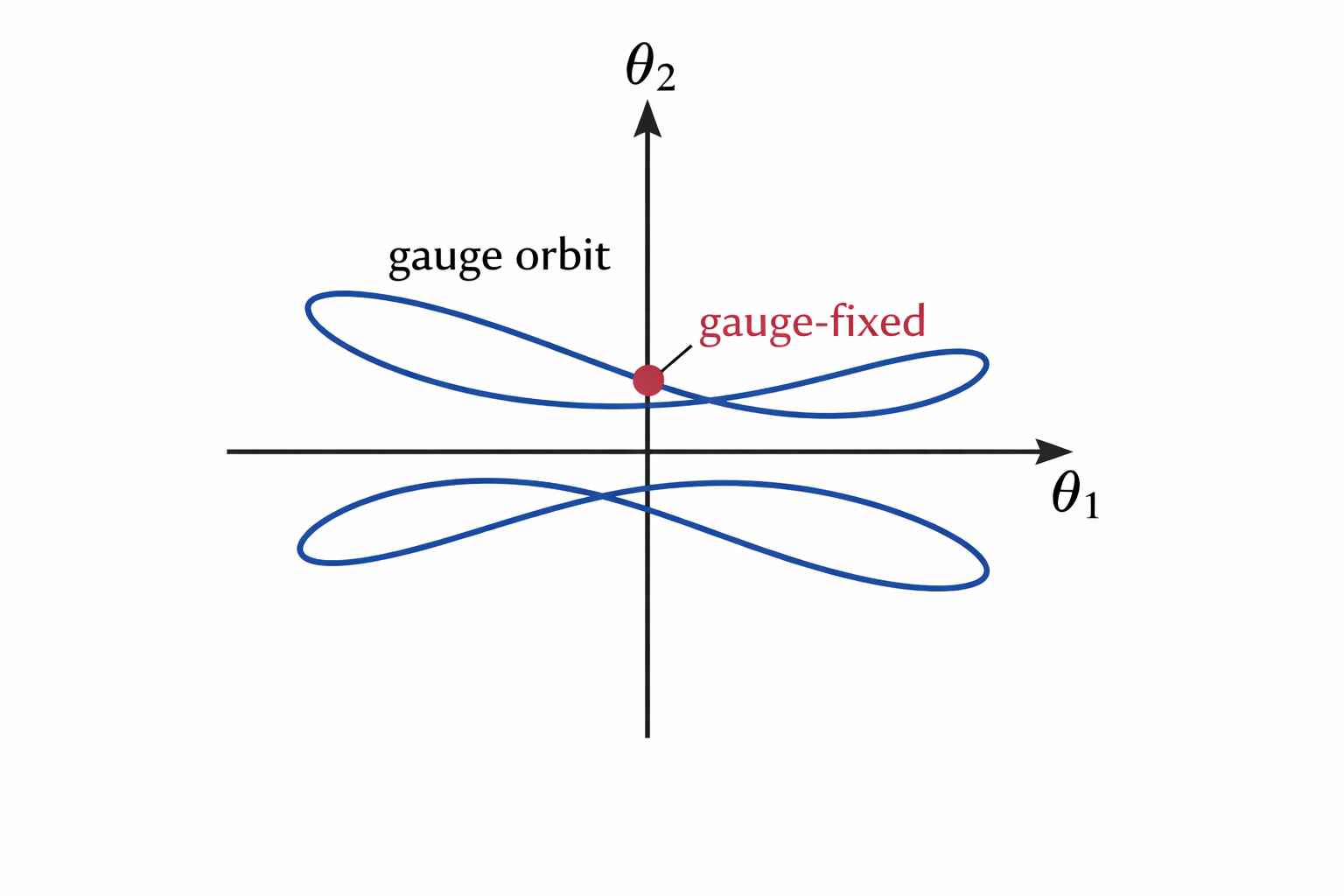}
\caption{
Gauge redundancy in parameter space.
Neuron-wise positive rescalings generate continuous gauge orbits.
}
\label{fig:gauge_orbit}
\end{figure}

In particular, gradient-based methods do not uniquely select a representative
within each equivalence class.
Parameters may therefore drift toward strongly imbalanced configurations
in which adjacent layers differ significantly in scale,
even though the realized function remains identical.
This observation motivates the introduction of a gauge-fixing condition,
in close analogy with gauge fixing in field theory,
to remove the redundant directions and modify the geometry of the
parameter space explored during training.
The structure of this redundancy is illustrated schematically in
Fig.~\ref{fig:gauge_orbit}, where different parameter configurations
connected by neuron-wise rescaling correspond to the same network
function.
The Figure~\ref{fig:gauge_orbit} above summarizes the central geometric picture:
because of ReLU homogeneity, neuron-wise rescalings move parameters along
continuous orbits that leave $f(x)$ invariant.
Optimization performed directly in parameter space can drift along these flat
directions, changing internal scales without functional benefit.
Gauge fixing introduces a preferred slice intersecting each orbit, thereby
lifting the redundant directions in a controlled way.

\section{Gauge Fixing by Norm Balancing}
\label{sec:gauge_fixing}

Having identified an exact reparametrization (gauge) redundancy in
Sec.~\ref{sec:gauge_symmetry}, we now introduce a concrete gauge-fixing prescription.
Our goal is to select, within each equivalence class of parameters representing the
same function, a canonical representative that avoids extreme internal
rescalings while leaving the realized mapping unchanged.

For the one-hidden-layer ReLU network in Eq.~\eqref{eq:relu_net}, the gauge
transformation~\eqref{eq:gauge_transform} rescales each hidden neuron
independently:
\begin{equation}
z(x) = W_1 x + b_1 \in \mathbb{R}^{H}, \qquad
h(x) = \sigma(z(x)) \in \mathbb{R}^{H},
\end{equation}
as
\begin{equation}
z(x)\to D z(x), \qquad h(x)\to D h(x),
\end{equation}
while compensating this rescaling by transforming the outgoing weights
$W_2 \to W_2 D^{-1}$.
Although $f(x)$ remains invariant, the internal parametrization can become
strongly imbalanced:
hidden activations may acquire large magnitude while the corresponding
outgoing weights become small, or vice versa.

To remove this degeneracy we impose a balancing condition between the
incoming and outgoing weights of each hidden neuron.
Let $W_{1,i}\in\mathbb{R}^{d}$ denote the $i$-th row of $W_1$
(incoming weights of neuron $i$), and let $W_{2,:,i}\in\mathbb{R}^{m}$
denote the corresponding column of $W_2$
(outgoing weights of neuron $i$).
We define the Euclidean norms
\begin{equation}
n_{1,i} \equiv \|W_{1,i}\|_2, \qquad
n_{2,i} \equiv \|W_{2,:,i}\|_2.
\end{equation}

Under the gauge transformation~\eqref{eq:gauge_transform},
\begin{equation}
n_{1,i} \to s_i\, n_{1,i}, \qquad
n_{2,i} \to \frac{1}{s_i}\, n_{2,i},
\label{eq:norm_transform}
\end{equation}
so that the product $n_{1,i} n_{2,i}$ is invariant while the ratio
$n_{1,i}/n_{2,i}$ can be tuned arbitrarily.
A natural gauge condition therefore consists in fixing this ratio.

Rather than imposing a hard constraint, we introduce a soft
gauge-fixing functional that penalizes scale imbalance multiplicatively:
\begin{equation}
\mathcal{G}(\theta)
=
\frac{1}{H}\sum_{i=1}^{H}
\left[
\log(n_{1,i}+\varepsilon) - \log(n_{2,i}+\varepsilon)
\right]^2 ,
\label{eq:gauge_functional}
\end{equation}
where $\theta$ denotes all trainable parameters and
$\varepsilon>0$ is a small constant introduced to ensure numerical stability
when norms approach zero.

The functional~\eqref{eq:gauge_functional} vanishes at perfectly balanced
representatives. In fact, each rescaling orbit admits (at least) one balanced
representative, obtained by an explicit neuron-wise rescaling:
Assume $n_{1,i},n_{2,i}>0$ for all hidden units. Consider the neuron-wise rescaling
$w_i\mapsto s_i w_i$, $a_i\mapsto s_i^{-1}a_i$ with $s_i>0$.
For $\varepsilon=0$, choosing $s_i^\star=\sqrt{n_{2,i}/n_{1,i}}$ yields
$n'_{1,i}=n'_{2,i}=\sqrt{n_{1,i}n_{2,i}}$ and therefore $\mathcal{G}(\theta')=0$.
For $\varepsilon>0$, the choice $s_i^\star=\sqrt{(n_{2,i}+\varepsilon)/(n_{1,i}+\varepsilon)}$
minimizes $\mathcal{G}$ along the orbit up to $O(\varepsilon)$.
Training is performed by minimizing the modified objective
\begin{equation}
\mathcal{L}_{\mathrm{total}}(\theta)
=
\mathcal{L}_{\mathrm{task}}(\theta)
+
\lambda\, \mathcal{G}(\theta),
\label{eq:total_loss}
\end{equation}
where $\mathcal{L}_{\mathrm{task}}$ is the task loss
(mean-squared error in our experiments)
and $\lambda\ge 0$ controls the strength of gauge fixing. We emphasize that $\lambda$ is not a gauge parameter in the
field-theoretic sense.
In gauge theory, changing a gauge parameter corresponds to selecting
different representatives of the same physical configuration
without changing physical predictions.
Here, by contrast, $\lambda$ is a penalty weight that modifies the
optimization objective and therefore the training dynamics.
The gauge structure refers solely to the exact rescaling equivalence
$\theta \sim \theta'$ induced by positive homogeneity.
The functional $\mathcal{G}$ selects a representative within each orbit,
whereas $\lambda$ controls how strongly this selection is represented
during training.

The parameter $\lambda$ interpolates between:
\begin{itemize}
  \item $\lambda=0$: full gauge redundancy; optimization may drift along
        flat directions.
  \item small $\lambda$: weak lifting of the degeneracy, modifying the
        geometry of parameter space while leaving the task objective dominant.
  \item large $\lambda$: strong enforcement of scale balance, which may
        interfere with early optimization transients.
\end{itemize}

It is important to distinguish this construction from common normalization
techniques.
Batch normalization modifies activations and changes the forward pass.
Weight decay penalizes large norms globally.
In contrast, the present gauge-fixing term:
\begin{itemize}
  \item acts exclusively on redundant scale directions,
  \item leaves the network function invariant under
        Eq.~\eqref{eq:gauge_transform},
  \item targets relative imbalance rather than absolute magnitude.
\end{itemize}

In this sense, Eq.~\eqref{eq:gauge_functional} plays a role analogous to
gauge fixing in field theory: it selects a canonical representative within
each equivalence class of parametrizations without modifying the physical
content of the model.

Because the transformation~\eqref{eq:gauge_transform} follows directly from
the positive homogeneity of ReLU, it provides an exact diagnostic of the
reparametrization symmetry.
For any set of strictly positive scaling factors $\{s_i\}$, the transformed
parameters $\theta'$ represent the same input--output function as the original
parameters $\theta$.
To verify this explicitly, after training we sample random positive scaling
factors and apply the transformation~\eqref{eq:gauge_transform} to the learned
parameters.
We then compute the mean absolute deviation between the original and transformed
network outputs,
\begin{equation}
\Delta_{\mathrm{inv}}
\equiv
\left\langle
\left|
f_{\theta}(x) - f_{\theta'}(x)
\right|
\right\rangle_x ,
\label{eq:invariance_error}
\end{equation}
where the average is taken over a test set.

In our numerical experiments performed in double precision (NumPy default),
$\Delta_{\mathrm{inv}}$ remains at the level of machine accuracy.
Across $200$ random gauge transformations we obtain
\[
\begin{aligned}
\min \Delta_{\mathrm{inv}} &= 3.99\times10^{-18}, \\
\mathrm{median} &= 6.68\times10^{-18}, \\
\max &= 1.14\times10^{-17}.
\end{aligned}
\]
These values are consistent with floating-point round-off error and confirm
that the gauge symmetry is exact for the class of networks considered.

\section{Gradient-flow dynamics in gauge coordinates}
\label{sec:gauge_flow}

The neuron-wise rescaling symmetry induced by positive homogeneity implies the
existence of flat directions in parameter space.
In this section we make this structure explicit at the level of training
dynamics by deriving the gradient-flow equations in coordinates adapted to the
gauge orbits.
The result is a simple decomposition: the task loss acts as a driving force that
can generate scale drift, while the gauge-fixing term induces an explicit
dissipative relaxation of the gauge (imbalance) modes.

For the $i$-th hidden neuron we denote the incoming and outgoing weight blocks by
\begin{equation}
w_i \equiv W_{1,i}\in\mathbb{R}^d,
\qquad
a_i \equiv W_{2,:,i}\in\mathbb{R}^m,
\end{equation}
with Euclidean norms
\begin{equation}
n_{1,i}\equiv \|w_i\|_2,\qquad n_{2,i}\equiv \|a_i\|_2.
\end{equation}
We define logarithmic norm coordinates (with $\varepsilon>0$ for numerical stability)
\begin{equation}
\alpha_i \equiv \log(n_{1,i}+\varepsilon),\qquad
\beta_i  \equiv \log(n_{2,i}+\varepsilon),
\end{equation}
and their sum/difference
\begin{equation}
u_i \equiv \alpha_i+\beta_i,\qquad
v_i \equiv \alpha_i-\beta_i.
\label{eq:u_v_def}
\end{equation}
Under neuron-wise rescaling $w_i\to s_i w_i$, $a_i\to s_i^{-1}a_i$ ($s_i>0$),
one has $u_i$ invariant while $v_i\to v_i+2\log s_i$.
Thus $v_i$ parametrizes the gauge (imbalance) direction along the orbit, whereas
$u_i$ is gauge-invariant.
Training with the soft gauge fixing
\begin{equation}
\mathcal{L}_{\mathrm{tot}}(\theta)=\mathcal{L}_{\mathrm{task}}(\theta)
+\lambda\,\mathcal{G}(\theta),
\qquad
\mathcal{G}(\theta)=\frac{1}{H}\sum_{i=1}^H v_i^2
\label{eq:gauge_fix_term}
\end{equation}
induces the gradient-flow dynamics
\begin{equation}
\dot{w}_i=-\nabla_{w_i}\mathcal{L}_{\mathrm{task}}-\lambda\nabla_{w_i}\mathcal{G},
\qquad
\dot{a}_i=-\nabla_{a_i}\mathcal{L}_{\mathrm{task}}-\lambda\nabla_{a_i}\mathcal{G},
\label{eq:gf_flow_blocks}
\end{equation}
where dot denotes derivative with respect to the (continuous) training time.
A direct computation (given in App.~\ref{app:gauge_flow_details}) yields
\begin{equation}
\nabla_{w_i}\mathcal{G}=\frac{2}{H}v_i\,
\frac{w_i}{n_{1,i}(n_{1,i}+\varepsilon)},
\qquad
\nabla_{a_i}\mathcal{G}=-\frac{2}{H}v_i\,
\frac{a_i}{n_{2,i}(n_{2,i}+\varepsilon)}.
\label{eq:grad_G_blocks}
\end{equation}
Therefore the gauge-fixing contribution is purely radial in each block:
it changes norms without rotating directions in weight space.

The induced dynamics for the gauge coordinate $v_i$ can be written in closed form.
Define the task-induced radial forces
\begin{equation}
F^{(1)}_i \equiv \left\langle \nabla_{w_i}\alpha_i,\nabla_{w_i}\mathcal{L}_{\mathrm{task}}\right\rangle,
\qquad
F^{(2)}_i \equiv \left\langle \nabla_{a_i}\beta_i,\nabla_{a_i}\mathcal{L}_{\mathrm{task}}\right\rangle,
\label{eq:radial_forces}
\end{equation}
where $\langle\cdot,\cdot\rangle$ denotes the Euclidean inner product.
Then one obtains (App.~\ref{app:gauge_flow_details})
\begin{equation}
\dot{v}_i
=
-\Big(F^{(1)}_i-F^{(2)}_i\Big)
-\frac{2\lambda}{H}\,v_i\left[
\frac{1}{(n_{1,i}+\varepsilon)^2}+\frac{1}{(n_{2,i}+\varepsilon)^2}
\right].
\label{eq:v_flow_full}
\end{equation}
The first term acts as a driving force generated by the task loss and may
produce scale drift along gauge orbits.
The second term is strictly dissipative and provides an explicit damping rate
\begin{equation}
\kappa_i\equiv
\frac{2\lambda}{H}\left[
\frac{1}{(n_{1,i}+\varepsilon)^2}+\frac{1}{(n_{2,i}+\varepsilon)^2}
\right]>0,
\qquad
\dot{v}_i\big|_{\mathrm{g}}=-\kappa_i v_i.
\label{eq:kappa_def}
\end{equation}
In particular, in the absence of task forcing ($F^{(1)}_i=F^{(2)}_i$) the imbalance
coordinate relaxes exponentially along each orbit,
\begin{equation}
v_i(t)=v_i(0)\exp\!\left(-\int_0^t \kappa_i(\tau)\,d\tau\right).
\label{eq:v_relax_general}
\end{equation}
Near the balanced regime $n_{1,i}\approx n_{2,i}\approx n_i$ one has the estimate
\begin{equation}
\kappa_i \approx \frac{4\lambda}{H}\frac{1}{(n_i+\varepsilon)^2},
\qquad
v_i(t)\approx v_i(0)\exp\!\left(-\frac{4\lambda}{H}\frac{t}{(n_i+\varepsilon)^2}\right).
\label{eq:v_relax_balanced}
\end{equation}
Equation~\eqref{eq:v_flow_full} makes precise the mechanism behind the observed
suppression of scale drift: the gauge-fixing term adds an explicit dissipative
potential along the gauge coordinates while leaving the input--output function invariant.

\section{Experiments}
\label{sec:experiments}

We now assess the practical implications of the gauge-fixing term introduced
in Sec.~\ref{sec:gauge_fixing}.
The experiments are deliberately minimal in order to isolate the structural
effects of reparametrization symmetry, rather than to optimize task performance.

We consider a one-dimensional nonlinear regression task with target function
\begin{equation}
y(x) = \sin(2.5x) + 0.2\cos(6x) + 0.1x + \eta,
\end{equation}
where $\eta$ is Gaussian noise with fixed variance.
Training samples are drawn uniformly from $[-2,2]$,
while validation is performed on a wider interval $[-3,3]$.
\begin{figure}[tbp]
\centering
\includegraphics[width=0.55\linewidth]{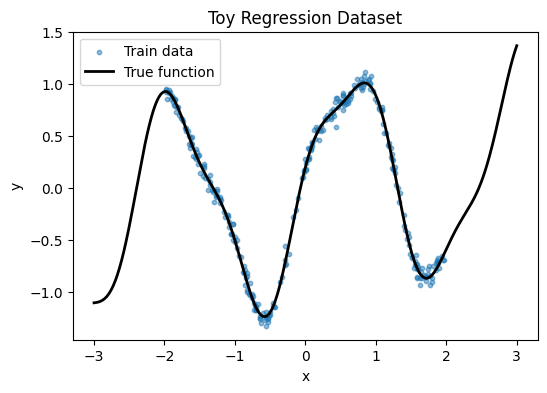}
\caption{
Toy regression dataset used in Sec.~\ref{sec:experiments}.
Training samples are drawn from $[-2,2]$ with additive Gaussian noise.
The black curve shows the noiseless target function evaluated on the validation grid
over $[-3,3]$.
}
\label{fig:dataset}
\end{figure}
Figure~\ref{fig:dataset} illustrates the controlled regression setting.
The training set is intentionally noisy, while validation is performed on a wider
interval to probe generalization.
This minimal setup is sufficient to expose optimization effects associated with
reparametrization redundancy without conflating them with large-model capacity.
All experiments use the single-hidden-layer ReLU network of
Eq.~\eqref{eq:relu_net} with width $H=20$.
The total loss \eqref{eq:total_loss} is minimized using plain gradient descent
with fixed learning rate.
Each configuration is trained for a fixed number of steps and averaged over
multiple random seeds.

\begin{figure}
\centering
\includegraphics[width=0.55\linewidth]{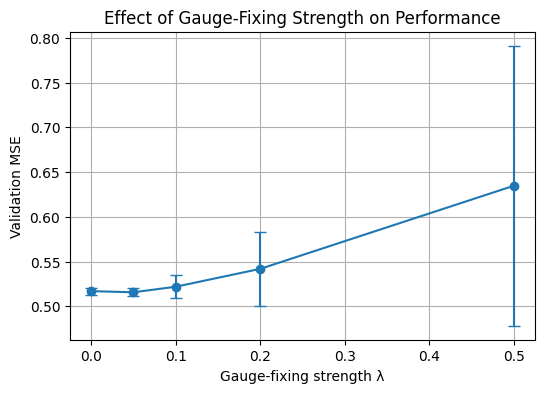}
\caption{
Validation mean-squared error as a function of the gauge-fixing strength $\lambda$.
Each point corresponds to the mean over independent random seeds, with error bars
given by the standard deviation.
}
\label{fig:lambda_sweep}
\end{figure}

We first investigate the effect of the gauge-fixing strength $\lambda$.
Figure~\ref{fig:lambda_sweep} and Table~\ref{tab:main_results} report the mean and
standard deviation of training and validation mean-squared error (MSE) for several
values of $\lambda$.
Figure~\ref{fig:lambda_sweep} shows that weak gauge fixing ($\lambda \approx 0.05$)
yields validation performance comparable to the baseline, whereas larger values of
$\lambda$ increase both the mean validation error and its variability across seeds.
Quantitatively, Table~\ref{tab:main_results} shows a mild optimum near $\lambda=0.05$
($0.5156 \pm 0.0044$) compared to the baseline $\lambda=0$ ($0.5169 \pm 0.0038$),
while strong gauge fixing substantially degrades generalization
(e.g.\ $\lambda=0.50$: $0.63 \pm 0.16$).
This supports the interpretation of $\mathcal{G}(\theta)$ as a geometric conditioning term:
useful when weak, but overly restrictive when too strong in this minimal setup.
\begin{table}
\centering
\begin{tabular}{lcc}
\hline\hline
$\lambda$ & Train MSE & Validation MSE \\
\hline
0.00 & $0.5358 \pm 0.0017$ & $0.5169 \pm 0.0038$ \\
0.05 & $0.5312 \pm 0.0025$ & $0.5156 \pm 0.0044$ \\
0.10 & $0.5281 \pm 0.0055$ & $0.5218 \pm 0.0128$ \\
0.20 & $0.5240 \pm 0.0083$ & $0.5418 \pm 0.0413$ \\
0.50 & $0.5177 \pm 0.0081$ & $0.63 \pm 0.16$ \\
\hline\hline
\end{tabular}
\caption{
Mean $\pm$ standard deviation over independent random seeds.
}
\label{tab:main_results}
\end{table}
To further probe stability, we vary the learning rate while comparing
baseline and gauge-fixed training at $\lambda=0.2$.
Table~\ref{tab:lr_stress} reports the validation MSE together with a qualitative
stability label.

\begin{table}
\centering
\begin{tabular}{lccc}
\hline\hline
Method & Learning rate & Validation MSE & Stability \\
\hline
Baseline & $5\times10^{-3}$ & $0.515360$ & Stable \\
Baseline & $1\times10^{-2}$ & $0.516924$ & Marginal \\
Baseline & $2\times10^{-2}$ & $0.521469$ & Unstable \\[0.3em]
Gauge-fixed ($\lambda=0.2$) & $5\times10^{-3}$ & $0.524778$ & Stable \\
Gauge-fixed ($\lambda=0.2$) & $1\times10^{-2}$ & $0.541801$ & Stable \\
Gauge-fixed ($\lambda=0.2$) & $2\times10^{-2}$ & $0.567897$ & Stable \\
\hline\hline
\end{tabular}
\caption{
Learning-rate stress test.
Gauge-fixed training remains stable at learning rates where baseline training
becomes unstable in this toy setup.
}
\label{tab:lr_stress}
\end{table}

At $\mathrm{lr}=2\times10^{-2}$, baseline training becomes unstable across seeds,
while the gauge-fixed configuration remains stable.
However, stability does not necessarily imply lower validation error at aggressive
learning rates: for $\lambda=0.2$ the stable runs exhibit larger validation MSE
than baseline training at smaller learning rates.
This indicates that gauge fixing modifies the stability region of gradient descent,
while the value of $\lambda$ controls the trade-off between constraint strength and
task optimization.

\begin{figure}
\centering
\includegraphics[width=0.55\linewidth]{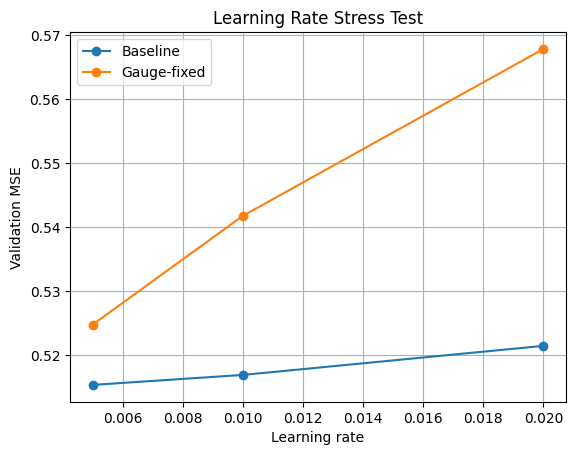}
\caption{
Learning-rate stress test comparing baseline and gauge-fixed training ($\lambda=0.2$).
Gauge fixing expands the empirically stable learning-rate regime in this toy setup.
}
\label{fig:lr_stress}
\end{figure}

\begin{figure}
\centering
\includegraphics[width=0.42\linewidth]{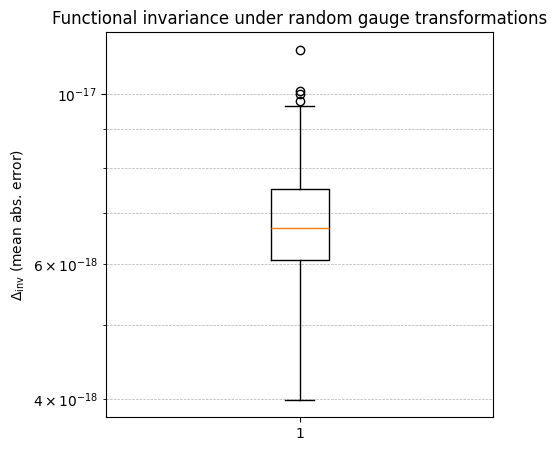}
\caption{
Functional invariance under random gauge transformations.
Shown is a boxplot of the invariance error $\Delta_{\mathrm{inv}}$
defined in Eq.~\eqref{eq:invariance_error} across 200 random neuron-wise
rescalings with $s_i>0$.
All values remain at double-precision numerical accuracy
(min $3.99\times 10^{-18}$, median $6.68\times 10^{-18}$, max $1.14\times 10^{-17}$),
confirming that the gauge transformations leave the network function unchanged
up to machine precision.
}
\label{fig:invariance}
\end{figure}

\FloatBarrier
As a final consistency check, we apply random gauge transformations to trained
models and compute the invariance error $\Delta_{\mathrm{inv}}$
defined in Eq.~\eqref{eq:invariance_error}.
The resulting distribution, shown in Fig.~\ref{fig:invariance},
remains at double-precision numerical accuracy ($\mathcal{O}(10^{-17})$),
confirming that gauge fixing does not alter the realized function.

\FloatBarrier
\section{Discussion and Interdisciplinary Perspective}
\label{sec:discussion}

The experiments of Sec.~\ref{sec:experiments} confirm that neural networks with
positively homogeneous activations possess an exact continuous redundancy in
parameter space and that softly lifting this redundancy modifies optimization
dynamics without altering the realized function.
Beyond the immediate numerical observations, this structure has conceptual
implications in both directions: for optimization theory in computer science
and for structural reasoning inspired by gauge theory in physics.

From the viewpoint of machine learning, neuron-wise rescaling invariance
implies that the loss landscape contains continuous flat directions
corresponding to redundant parametrizations.
These directions enlarge the region explored by gradient descent without
changing predictive performance.
The imbalance coordinates $v_i$ introduced in Sec.~\ref{sec:gauge_flow}
make these redundant modes explicit.
In this coordinate system, the gauge-fixing term adds a strictly dissipative
contribution $\dot v_i|_{\mathrm g}=-\kappa_i v_i$,
revealing that part of the instability observed at large learning rates
originates from unconstrained drift along symmetry directions rather than from
model capacity or task complexity.
In this sense, gauge fixing acts as a geometric conditioning mechanism:
it reshapes the effective optimization geometry without modifying the model’s
expressive power.

From the viewpoint of theoretical physics, the correspondence is structural
rather than metaphorical.
The symmetry arises from an exact positive homogeneity property and generates
continuous orbits in parameter space.
Gauge-invariant combinations (such as $u_i$) remain untouched,
while redundant coordinates ($v_i$) can be isolated and controlled.
The introduction of $\mathcal{G}(\theta)$ parallels gauge fixing in field theory:
it selects a representative on each orbit while preserving physical observables.
Here the ``observables'' are the input--output functions $f(x)$.
This establishes a concrete example in which gauge-theoretic language
provides a mathematically precise description of optimization geometry
outside traditional high-energy physics contexts.

The interdisciplinary value lies in the bidirectional transfer of structure.
On the machine-learning side, orbit–slice decompositions clarify how
reparametrization symmetries affect conditioning and stability.
On the physics side, neural networks provide an accessible setting in which
gauge redundancy, orbit geometry, and soft symmetry lifting can be studied
in finite-dimensional systems with explicit dynamics.
This interplay suggests that methods developed in one domain may inform the
other: quotient-space optimization techniques in ML resemble gauge-fixed
formulations in physics, while renormalization-inspired reasoning about
scale flow may illuminate stability properties in deep networks.

Importantly, the primary role of $\mathcal{G}(\theta)$ is geometric rather
than regularizing.
Weak gauge fixing modifies the stability boundary of gradient descent
while preserving validation performance,
whereas strong gauge fixing constrains the dynamics excessively.
This trade-off reflects a familiar principle from gauge theory:
fixing redundant directions improves mathematical control,
but overconstraining can obscure useful dynamical freedom.

Although the present experiments are deliberately minimal,
the underlying symmetry mechanism is general.
Any architecture with positively homogeneous activations exhibits analogous
scale redundancy.
In deeper networks these symmetries propagate across layers,
leading to richer orbit structures.
Understanding these redundancies in geometric terms may offer a principled
framework for analyzing conditioning, learning-rate stability,
and parameter-space curvature in modern architectures.

Generally our work shows that gauge-theoretic reasoning
can function as a structural organizing principle in optimization theory.
The symmetry is exact, the orbits are explicit,
the gauge coordinates are computable,
and the gauge-fixing term has a direct dynamical interpretation.
This suggests that further development of symmetry-adapted coordinates,
orbit-based diagnostics, and controlled symmetry lifting
may provide a systematic bridge between high-energy physics concepts
and large-scale learning systems.

\section{Conclusion}
\label{sec:conclusion}

We have identified an exact continuous reparametrization symmetry
in neural networks with positively homogeneous activations
and interpreted it as a gauge redundancy in parameter space.
This symmetry generates flat directions along which the loss remains invariant,
allowing internal scale redistribution without altering the realized function.

Inspired by gauge fixing in field theory,
we introduced a norm-balancing functional that softly lifts this degeneracy
while preserving the input--output mapping.
The proposed gauge-fixing term does not restrict expressivity;
rather, it selects a canonical representative within each equivalence class
of parametrizations.

In a minimal regression experiment,
we observed that weak gauge fixing leaves performance essentially unchanged,
while strong gauge fixing can degrade validation error.
However, the gauge-fixed configuration remains stable in regimes
where baseline gradient descent becomes unstable,
indicating that lifting redundant scale directions
modifies optimization geometry in a nontrivial way.

These results suggest that part of the instability observed in
homogeneous neural networks may be attributed to intrinsic
reparametrization redundancy rather than solely to model capacity
or optimizer choice.
Controlling such redundancies provides a principled mechanism
for reshaping the effective parameter landscape explored during training.

More broadly, this work shows that the analogy between neural
network reparametrization and gauge symmetry can be made precise and operational.
The symmetry is exact, the gauge orbits are explicit,
and the gauge-fixing functional has a direct computational realization.
Extending this perspective to deeper architectures
and to other classes of homogeneous networks
may offer further insight into the geometry of modern learning systems.

Lastly, here we address a parameter-space gauge redundancy induced by positive homogeneity:
neuron-wise rescalings generate continuous orbits along which the network function is exactly invariant.
Our norm-balancing term acts as a soft gauge fixing, selecting a canonical representative within each orbit
and modifying optimization geometry without changing expressivity.
This is complementary to a dynamical gauge invariance, as realized in our GINN framework,
where Ward identities constrain fluctuation corrections and protect the critical stability boundary.
A natural extension is to embed the present construction into a stochastic QFT-style effective action
and promote architectures to slow variables evolving under a Markovian (Lindblad-type) dynamics in function space,
thereby unifying parameter-space gauge fixing with symmetry-protected stochastic evolution.

\appendix
\section{Derivation of the gradient-flow equations in gauge coordinates}
\label{app:gauge_flow_details}

In this appendix we provide the detailed derivation of the gradient-flow dynamics
in the gauge-adapted coordinates introduced in Sec.~\ref{sec:gauge_flow}.

\subsection{Preliminaries and notation}

For each hidden neuron $i$ we denote
\begin{equation}
w_i \equiv W_{1,i}\in\mathbb{R}^d,
\qquad
a_i \equiv W_{2,:,i}\in\mathbb{R}^m,
\end{equation}
and their norms
\begin{equation}
n_{1,i}\equiv \|w_i\|_2=\sqrt{w_i^\top w_i},
\qquad
n_{2,i}\equiv \|a_i\|_2=\sqrt{a_i^\top a_i}.
\end{equation}
We define
\begin{equation}
\alpha_i \equiv \log(n_{1,i}+\varepsilon),
\qquad
\beta_i \equiv \log(n_{2,i}+\varepsilon),
\qquad
v_i \equiv \alpha_i-\beta_i,
\qquad
u_i \equiv \alpha_i+\beta_i,
\end{equation}
with $\varepsilon>0$.

The gauge-fixing functional is
\begin{equation}
\mathcal{G}(\theta)=\frac{1}{H}\sum_{i=1}^H v_i^2,
\qquad
\mathcal{L}_{\mathrm{tot}}=\mathcal{L}_{\mathrm{task}}+\lambda\mathcal{G}.
\end{equation}
Gradient flow is defined by
\begin{equation}
\dot{\theta}=-\nabla_\theta \mathcal{L}_{\mathrm{tot}}.
\end{equation}

\subsection{Gradients of the logarithmic norm coordinates}

We first compute $\nabla_{w_i}\alpha_i$.
Using $\alpha_i=\log(n_{1,i}+\varepsilon)$ and the chain rule,
\begin{equation}
\nabla_{w_i}\alpha_i
=
\frac{1}{n_{1,i}+\varepsilon}\,\nabla_{w_i}n_{1,i}.
\label{eq:chain_alpha}
\end{equation}
Since $n_{1,i}=\sqrt{w_i^\top w_i}$, one has
\begin{equation}
\nabla_{w_i}n_{1,i}
=
\nabla_{w_i}(w_i^\top w_i)^{1/2}
=
\frac{1}{2}(w_i^\top w_i)^{-1/2}\,\nabla_{w_i}(w_i^\top w_i)
=
\frac{1}{2n_{1,i}}\,2w_i
=
\frac{w_i}{n_{1,i}}.
\label{eq:grad_norm_w}
\end{equation}
Substituting~\eqref{eq:grad_norm_w} into~\eqref{eq:chain_alpha} yields
\begin{equation}
\nabla_{w_i}\alpha_i
=
\frac{w_i}{n_{1,i}(n_{1,i}+\varepsilon)}.
\label{eq:grad_alpha}
\end{equation}
Similarly,
\begin{equation}
\nabla_{a_i}\beta_i
=
\frac{a_i}{n_{2,i}(n_{2,i}+\varepsilon)}.
\label{eq:grad_beta}
\end{equation}

Because $v_i=\alpha_i-\beta_i$, and $\alpha_i$ depends only on $w_i$ while $\beta_i$
depends only on $a_i$, we obtain
\begin{equation}
\nabla_{w_i}v_i=\nabla_{w_i}\alpha_i
=
\frac{w_i}{n_{1,i}(n_{1,i}+\varepsilon)},
\qquad
\nabla_{a_i}v_i=-\nabla_{a_i}\beta_i
=
-\frac{a_i}{n_{2,i}(n_{2,i}+\varepsilon)}.
\label{eq:grad_v_blocks}
\end{equation}

\subsection{Gradients of the gauge-fixing functional}

Since $\mathcal{G}=\frac{1}{H}\sum_j v_j^2$ and $v_j$ depends on $(w_j,a_j)$ only,
\begin{equation}
\nabla_{w_i}\mathcal{G}
=
\frac{1}{H}\nabla_{w_i}(v_i^2)
=
\frac{2}{H}v_i\,\nabla_{w_i}v_i
=
\frac{2}{H}v_i\,
\frac{w_i}{n_{1,i}(n_{1,i}+\varepsilon)}.
\label{eq:gradG_w}
\end{equation}
Likewise,
\begin{equation}
\nabla_{a_i}\mathcal{G}
=
\frac{1}{H}\nabla_{a_i}(v_i^2)
=
\frac{2}{H}v_i\,\nabla_{a_i}v_i
=
-\frac{2}{H}v_i\,
\frac{a_i}{n_{2,i}(n_{2,i}+\varepsilon)}.
\label{eq:gradG_a}
\end{equation}
Equations~\eqref{eq:gradG_w}--\eqref{eq:gradG_a} show that the gauge term acts purely
radially in each weight block.

\subsection{Gauge-only contribution to the block dynamics}

Separating task and gauge parts in gradient flow,
\begin{equation}
\dot{w}_i=-\nabla_{w_i}\mathcal{L}_{\mathrm{task}}-\lambda\nabla_{w_i}\mathcal{G},
\qquad
\dot{a}_i=-\nabla_{a_i}\mathcal{L}_{\mathrm{task}}-\lambda\nabla_{a_i}\mathcal{G},
\end{equation}
the gauge-only components are obtained by dropping $\mathcal{L}_{\mathrm{task}}$:
\begin{equation}
\dot{w}_i\big|_{\mathrm{g}}
=-\lambda\,\nabla_{w_i}\mathcal{G}
=
-\lambda\frac{2}{H}v_i\,
\frac{w_i}{n_{1,i}(n_{1,i}+\varepsilon)},
\label{eq:w_g_only}
\end{equation}
\begin{equation}
\dot{a}_i\big|_{\mathrm{g}}
=-\lambda\,\nabla_{a_i}\mathcal{G}
=
+\lambda\frac{2}{H}v_i\,
\frac{a_i}{n_{2,i}(n_{2,i}+\varepsilon)}.
\label{eq:a_g_only}
\end{equation}

\subsection{Dynamics of $\alpha_i$ and $\beta_i$}

We now compute the time derivative of $\alpha_i$ under gradient flow.
By the chain rule,
\begin{equation}
\dot{\alpha}_i
=
\left\langle \nabla_{w_i}\alpha_i,\dot{w}_i\right\rangle.
\label{eq:alpha_dot_chain}
\end{equation}
For the gauge-only part, substitute~\eqref{eq:grad_alpha} and~\eqref{eq:w_g_only}:
\begin{align}
\dot{\alpha}_i\big|_{\mathrm{g}}
&=
\left\langle
\frac{w_i}{n_{1,i}(n_{1,i}+\varepsilon)},
-\lambda\frac{2}{H}v_i\,
\frac{w_i}{n_{1,i}(n_{1,i}+\varepsilon)}
\right\rangle
\nonumber\\
&=
-\lambda\frac{2}{H}v_i\,
\frac{\langle w_i,w_i\rangle}{n_{1,i}^2(n_{1,i}+\varepsilon)^2}
=
-\lambda\frac{2}{H}v_i\,
\frac{n_{1,i}^2}{n_{1,i}^2(n_{1,i}+\varepsilon)^2}
=
-\frac{2\lambda}{H}\frac{v_i}{(n_{1,i}+\varepsilon)^2}.
\label{eq:alpha_dot_g}
\end{align}
Similarly, using
\begin{equation}
\dot{\beta}_i
=
\left\langle \nabla_{a_i}\beta_i,\dot{a}_i\right\rangle,
\label{eq:beta_dot_chain}
\end{equation}
and substituting~\eqref{eq:grad_beta} and~\eqref{eq:a_g_only},
\begin{align}
\dot{\beta}_i\big|_{\mathrm{g}}
&=
\left\langle
\frac{a_i}{n_{2,i}(n_{2,i}+\varepsilon)},
+\lambda\frac{2}{H}v_i\,
\frac{a_i}{n_{2,i}(n_{2,i}+\varepsilon)}
\right\rangle
\nonumber\\
&=
+\lambda\frac{2}{H}v_i\,
\frac{\langle a_i,a_i\rangle}{n_{2,i}^2(n_{2,i}+\varepsilon)^2}
=
+\frac{2\lambda}{H}\frac{v_i}{(n_{2,i}+\varepsilon)^2}.
\label{eq:beta_dot_g}
\end{align}

\subsection{Closed dynamics for the gauge coordinate $v_i$}

Since $v_i=\alpha_i-\beta_i$, we obtain for the gauge-only part
\begin{align}
\dot{v}_i\big|_{\mathrm{g}}
&=
\dot{\alpha}_i\big|_{\mathrm{g}}-\dot{\beta}_i\big|_{\mathrm{g}}
\nonumber\\
&=
-\frac{2\lambda}{H}\frac{v_i}{(n_{1,i}+\varepsilon)^2}
-\frac{2\lambda}{H}\frac{v_i}{(n_{2,i}+\varepsilon)^2}
\nonumber\\
&=
-\frac{2\lambda}{H}\,v_i\left[
\frac{1}{(n_{1,i}+\varepsilon)^2}+\frac{1}{(n_{2,i}+\varepsilon)^2}
\right].
\label{eq:v_dot_g}
\end{align}
Defining the damping rate
\begin{equation}
\kappa_i\equiv
\frac{2\lambda}{H}\left[
\frac{1}{(n_{1,i}+\varepsilon)^2}+\frac{1}{(n_{2,i}+\varepsilon)^2}
\right]>0,
\end{equation}
we can write $\dot{v}_i\big|_{\mathrm{g}}=-\kappa_i v_i$.

Including the task loss, we introduce the task-induced radial forces
\begin{equation}
F^{(1)}_i \equiv \left\langle \nabla_{w_i}\alpha_i,\nabla_{w_i}\mathcal{L}_{\mathrm{task}}\right\rangle,
\qquad
F^{(2)}_i \equiv \left\langle \nabla_{a_i}\beta_i,\nabla_{a_i}\mathcal{L}_{\mathrm{task}}\right\rangle,
\end{equation}
so that
\begin{equation}
\dot{\alpha}_i = -F^{(1)}_i -\frac{2\lambda}{H}\frac{v_i}{(n_{1,i}+\varepsilon)^2},
\qquad
\dot{\beta}_i = -F^{(2)}_i +\frac{2\lambda}{H}\frac{v_i}{(n_{2,i}+\varepsilon)^2}.
\end{equation}
Subtracting yields the full imbalance dynamics
\begin{equation}
\dot{v}_i
=
-\Big(F^{(1)}_i-F^{(2)}_i\Big)
-\frac{2\lambda}{H}\,v_i\left[
\frac{1}{(n_{1,i}+\varepsilon)^2}+\frac{1}{(n_{2,i}+\varepsilon)^2}
\right],
\end{equation}
which is Eq.~( \ref{eq:v_flow_full} ) in the main text.

For completeness, we also record the gauge-only contribution to $u_i=\alpha_i+\beta_i$:
\begin{equation}
\dot{u}_i\big|_{\mathrm{g}}
=
\dot{\alpha}_i\big|_{\mathrm{g}}+\dot{\beta}_i\big|_{\mathrm{g}}
=
\frac{2\lambda}{H}v_i\left[
-\frac{1}{(n_{1,i}+\varepsilon)^2}+\frac{1}{(n_{2,i}+\varepsilon)^2}
\right].
\end{equation}
Near the balanced regime $n_{1,i}\approx n_{2,i}$ this term is subleading, so the
dominant effect of gauge fixing is dissipative relaxation of the gauge coordinate $v_i$.

\bibliographystyle{unsrtnat}
\bibliography{refs}

@article{CarmoTerin:2025lyx,
    author = "Carmo Terin, Rodrigo",
    title = "{The GINN framework: a stochastic QED correspondence for stability and chaos in deep neural networks}",
    eprint = "2508.18948",
    archivePrefix = "arXiv",
    primaryClass = "hep-th",
    month = "8",
    year = "2025"
}

@inproceedings{Yaida2020,
  author    = {Sho Yaida},
  title     = {Non-Gaussian Processes and Neural Networks at Finite Widths},
  booktitle = {Proceedings of Machine Learning Research},
  volume    = {107},
  pages     = {165--192},
  year      = {2020}
}

@inproceedings{DyerGurAri2020,
  author    = {Ethan Dyer and Guy Gur-Ari},
  title     = {Asymptotics of Wide Networks from Feynman Diagrams},
  booktitle = {International Conference on Learning Representations (ICLR)},
  year      = {2020},
  eprint    = {1909.11304},
  archivePrefix = {arXiv},
  primaryClass  = {cs.LG}
}

@inproceedings{BondesanWelling2021,
  author    = {Roberto Bondesan and Max Welling},
  title     = {The Hintons in your Neural Network: A Quantum Field Theory View of Deep Learning},
  booktitle = {Proceedings of the 38th International Conference on Machine Learning},
  volume    = {139},
  pages     = {1038--1048},
  year      = {2021}
}

@article{HalversonMaitiStoner2021,
  author  = {James Halverson and Anindita Maiti and Keegan Stoner},
  title   = {Neural Networks and Quantum Field Theory},
  journal = {Machine Learning: Science and Technology},
  volume  = {2},
  pages   = {035002},
  year    = {2021},
  doi     = {10.1088/2632-2153/abeca3}
}

@article{ErbinLahocheSamary2022,
  author  = {Harold Erbin and Vincent Lahoche and Dine Ousmane Samary},
  title   = {Nonperturbative Renormalization for the Neural Network--QFT Correspondence},
  journal = {Machine Learning: Science and Technology},
  volume  = {3},
  pages   = {015027},
  year    = {2022},
  doi     = {10.1088/2632-2153/ac4f69}
}

@book{Roberts2022,
  author    = {Daniel A. Roberts and Sho Yaida and Boris Hanin},
  title     = {The Principles of Deep Learning Theory},
  publisher = {Cambridge University Press},
  year      = {2022}
}

@article{GrosvenorJefferson2022,
  author  = {Kevin T. Grosvenor and Ro Jefferson},
  title   = {The Edge of Chaos: Quantum Field Theory and Deep Neural Networks},
  journal = {SciPost Physics},
  volume  = {12},
  pages   = {081},
  year    = {2022},
  doi     = {10.21468/SciPostPhys.12.3.081}
}

@article{Sompolinsky1988,
  author  = {H. Sompolinsky and A. Crisanti and H. J. Sommers},
  title   = {Chaos in Random Neural Networks},
  journal = {Physical Review Letters},
  volume  = {61},
  pages   = {259--262},
  year    = {1988},
  doi     = {10.1103/PhysRevLett.61.259}
}

@inproceedings{poole2016exponential,
  author    = {Ben Poole and Subhaneil Lahiri and Maithra Raghu and Jascha Sohl-Dickstein and Surya Ganguli},
  title     = {Exponential Expressivity in Deep Neural Networks through Transient Chaos},
  booktitle = {Advances in Neural Information Processing Systems},
  volume    = {29},
  pages     = {3360--3368},
  year      = {2016}
}

@article{Langton1990,
  author  = {C. G. Langton},
  title   = {Computation at the Edge of Chaos: Phase Transitions and Emergent Computation},
  journal = {Physica D},
  volume  = {42},
  pages   = {12--37},
  year    = {1990},
  doi     = {10.1016/0167-2789(90)90064-V}
}

@inproceedings{Meng2019GSGD,
  author    = {Qi Meng and Shuxin Zheng and Huishuai Zhang and Wei Chen and Qiwei Ye and Zhi-Ming Ma and Nenghai Yu and Tie-Yan Liu},
  title     = {$\mathcal{G}$-SGD: Optimizing ReLU Neural Networks in its Positively Scale-Invariant Space},
  booktitle = {International Conference on Learning Representations (ICLR)},
  year      = {2019}
}

@article{Saul2023WeightBalancing,
  author  = {Lawrence K. Saul},
  title   = {Weight-balancing Fixes and Flows for Deep Learning},
  journal = {Transactions on Machine Learning Research},
  year    = {2023}
}

@article{Brea2019WeightSymmetry,
  author  = {Johanni Brea and Berfin Simsek and Bernd Illing and Wulfram Gerstner},
  title   = {Weight-space Symmetry in Deep Networks Gives Rise to Permutation Saddles, Connected by Equal-loss Valleys across the Loss Landscape},
  journal = {arXiv preprint},
  year    = {2019},
  eprint  = {1907.02911},
  archivePrefix = {arXiv},
  primaryClass  = {cs.LG}
}
\end{document}